\documentclass[conference]{IEEEtran}
\IEEEoverridecommandlockouts
\usepackage{cite}
\usepackage{amsmath,amssymb,amsfonts}
\usepackage{algorithmic}
\usepackage{algorithm2e}
\usepackage{graphicx}
\usepackage{textcomp}
\usepackage{xcolor}
\def\BibTeX{{\rm B\kern-.05em{\sc i\kern-.025em b}\kern-.08em
    T\kern-.1667em\lower.7ex\hbox{E}\kern-.125emX}}

\begin{document}

\title{Improving Joint Layer RNN based Keyphrase Extraction by Using Syntactical Features}

\author{\IEEEauthorblockN{Miftahul Mahfuzh}
\IEEEauthorblockA{\textit{Prosa.ai}\\
Bandung, Indonesia \\
miftahul.mahfuzh@prosa.ai}
\and
\IEEEauthorblockN{Sidik Soleman}
\IEEEauthorblockA{\textit{Prosa.ai}\\
Bandung, Indonesia \\
sidik.soleman@prosa.ai}
\and
\IEEEauthorblockN{Ayu Purwarianti}
\IEEEauthorblockA{\textit{School of Electrical Engineering and Informatics} \\
\textit{Institut Teknologi Bandung}\\
Bandung, Indonesia \\
ayu@informatika.org}
}

\maketitle

\begin{abstract}
Keyphrase extraction as a task to identify important words or phrases from a text, is a crucial process to identify main topics when analyzing texts from a social media platform. In our study, we focus on text written in Indonesia language taken from Twitter. Different from the original joint layer recurrent neural network (JRNN) with output of one sequence of keywords and using only word embedding, here we propose to modify the input layer of JRNN to extract more than one sequence of keywords by additional information of syntactical features, namely part of speech, named entity types, and dependency structures. Since JRNN in general requires a large amount of data as the training examples and creating those examples is expensive, we used a data augmentation method to increase the number of training examples. Our experiment had shown that our method outperformed the baseline methods. Our method achieved $.9597$ in accuracy and $.7691$ in F1. 
\end{abstract}


\section{Introduction}
Through various social media platforms, people can express their opinions or complaints regarding some company products or services. Analyzing these texts is beneficial to the company since it gives various insights about the company products and services. Based on these insights, companies  could decide strategies in order to improve their qualities. 

Keyphrase extraction is one of the crucial tasks in analyzing texts from social medias. It aims to identify important words or phrases from the texts that describe their main ideas. Hence, the texts can be easily visualized for example by means of phrase cloud. Besides information visualization, there are some text processing tasks that can take advantage of keyphrase extraction such as information retrieval, automatic question answering, text classification, and text summarization. 

Different from keyword, keyphrase may contain more than one word. Keyphrase extraction is often being formulated as a ranking or a classification problem. In the ranking problem, phrases are assigned with some values that tell its degree of importance. The values are used to rank them and top-$N$ phrases are selected as keyphrases. While in the classification problem, the task is to decide whether a phrase is a keyphrase or not. In this paper, we formulated the keyphrase extraction as the classification problem. 

In this work, we focus on keyphrase extraction for texts from Twitter. The texts are usually short since Twitter only allows 280 characters for a single tweet. One of the current state-of-the-art methods is the one proposed by Zhang et al.~\cite{zhang}. Their method was based on joint layer neural recurrent neural network (JRNN) with word embedding as its input. It is a variant of a stacked RNN with two hidden layers. The outputs of both hidden layers are combined into an objective layer.

In their experiment, they assumed that a text has only a single keyphrase. We argue that a single keyphrase is not enough for representing the main ideas of a text. Therefore, we allowed for a text to have more than one keyphrase in our work. Unlike their research that targeted texts written in English, we targeted texts written in Indonesian language since little work has been done on it. 

Generally, a large amount of data is required for training when we use JRNN. However, creating those examples manually is expensive. To alleviate this problem, we used text augmentation method to increase the number of training examples. Furthermore, different with Zhang et al.~\cite{zhang}, not only word embedding feature, we also use additional features, namely part of speech, named entity types, and dependency structures. 

In this paper, our main contribution could be summarized as follows:
\begin{enumerate}
    \item To propose a method of extracting keyphrase using JRNN based architecture with multiple features. 
    \item To create a dataset (in Indonesian language) for the evaluation, and to propose a data augmentation method and use it to increase the number of training examples. 
    \item To evaluate the effectiveness of our method. Our experiment showed that our proposed method outperformed the state-of-the-art method.
\end{enumerate}

We have organized the rest of the paper in the following way. Section~\ref{sec:relatedwork} discusses the work related to ours and Section~\ref{sec:proposed} presents our proposed method. Section~\ref{sec:experiment} discusses our experimental results. In Section~\ref{sec:conclusion}, we conclude our work and discuss future work. 

\section{Related Work}
\label{sec:relatedwork}
In general, keyphrase extraction methods can be divided into two types, namely unsupervised and supervised methods. In the unsupervised methods, they do not need any training examples. Hence, ideally they should be domain independent. They usually formulate the task as a ranking problem. 

Mihalcea and Tarau~\cite{mihalcea} proposed a graph-based method called TextRank. Several words were selected from a given text. Two words were connected when they are in the same window of K words to create a graph. Score for each word was calculated based on its connections in the graph. Top-$N$ words were selected and sequences of adjacent words were merged into a phrase.

Since a phrase may not be extracted in TextRank because one of its words are not in the top-$N$ words, Rose et al.~\cite{rose} proposed RAKE that generated phrases before graph construction. Score for each phrase was calculated based on their relations in the graph and the frequencies of their word members. Top-$N$ phrases were selected as the keyphrases. 
 
Rini et al.~\cite{rini} proposed a method that assigned weights for phrases based on some heuristic weighting according to part of speech tags. They evaluated their method on extracting keyphrases related to complain from Twitter in Indonesia language. However, their dataset for the evaluation was very small about 50 texts.

Hasan and Ng~\cite{hasan} compared several unsupervised methods. They found that a method based on term frequency and inverted document frequency (TFxIDF) was very robust across different datasets.

Mahata et al.~\cite{debanjan} took advantage of phrase and document embedding. They gave score for each phrase based on its semantic similarity to the text and word co-occurrences. Bennani-Smires et al.~\cite{smires} also proposed similar method to Mahata et al.~\cite{debanjan} but the scores for phrases were calculated only based on the similarity score between the phrases and the text.

Most of the unsupervised methods above except the one proposed by Rini et al.~\cite{rini}, targeted long documents. Since most of them relied on word frequencies and co-occurrences, it is unlikely that the methods are applicable for short texts such as the ones from Twitter. 

In the supervised methods, they formulated the task as a classification problem, i.e. to classify whether a phrase is a keyphrase or not. Witten et al.~\cite{ian} used na\"ive bayes algorithm for keyphrase extraction. Zhang~\cite{zhangc} used condition random field to extract keyphrase from documents in chinese language. Thus, surrounding words could be considered to decide whether a phrase is a keyphrase. Zhang et al.~\cite{zhang} proposed a joint layer recurrent neural network (JRNN) for keyphrase extraction in Twitter (in English language). Their experiment showed a promising result.

So far there is not much work regarding keyphrase extraction from a single tweet in Twitter, especially in Indonesian language. Additionally, we were the first to use JRNN architecture for keyphrase extraction targeting text written in Indonesian language from Twitter.

\section{Proposed Method}
\label{sec:proposed}
In this work, we adopted the JRNN proposed by Zhang et al.~\cite{zhang}. In the next sub-section, we briefly explain their method and our modification. Then, we discuss data augmentation as one of the means to increase the amount of data, and our additional linguistic features used in our work, i.e. part of speech, named entity types and sentence dependency structure.

\subsection{Joint Layer RNN}
The Joint Layer RNN (JRNN) proposed by Zhang et al.~\cite{zhang} is an extension of the stacked RNN proposed by Pascanu et al.~\cite{pascanu}. Hence, the description and formulations of JRNN in this sub-section are based on Zhang et al.~\cite{zhang}. JRNN consists of two RNNs as hidden layers. Each RNN accepts the time series of previous layers as the input and transforms it into the output layers. Since there are two output layers, both layers are combined with linear combination in the objective function. 

The objective of the first RNN is to learn whether a word is an important one or not. This layer uses the input layer as its input. Its output prediction is ${\hat{y}}^{1}_{t}$ and its output target is ${y}^{1}_{t}$. 

The second RNN learns whether a word is a non-keyphrase, beginning of keyphrase, middle of keyphrase, end of keyphrase, or a keyphrase consisting of a single word. Hence, there are five classes for this layer. The output prediction and target of this layer is ${\hat{y}}^{2}_{t}$ and ${y}^{2}_{t}$, respectively. The input of this layer is the output of the first RNN. 

Different from Zhang et al.~\cite{zhang} with five classes for the second RNN, we proposed a simpler and fewer classes. We used three classes, i.e. non-keyphrase (labelled as 0), beginning of keyphrase (labelled as 1), and the tail of keyphrase (labelled as 2). Hence, words that are labelled as the middle and end of keyphrase are merged into the same class, i.e. the tail of keyphrase in our case. 

Here, we describe the formulations of JRNN based on the work proposed by Zhang et al.~\cite{zhang}. The following equation describes the formulation of the hidden layers:
\begin{equation}
    h^{1}_{t} = f_{hidden}(x_{t}, h^{1}_{t-1})
\end{equation}
\begin{equation}
     h^{2}_{t} = f_{hidden}(h^{1}_{t}, h^{2}_{t-1})
\end{equation}
where $x_t$, $h_t^1$, and $h_t^2$ is the training input, the first hidden layer, and the second hidden layer at time $t$. The output layer is calculated as follows:
\begin{equation}
     {\hat{y}}^{1}_{t} = f_{out}(h^{1}_{t})
\end{equation}
\begin{equation}
     {\hat{y}}^{2}_{t} =f_{out}(h^{2}_{t})
\end{equation}

Given N training examples, the total time T, and the parameter $\theta$ in JRNN, the objective or cost functions for each RNN are calculated as:
\begin{equation}
    J_1(\theta)=\frac{1}{N}\displaystyle \sum_{n=1}^{N} \sum_{t=1}^{t=T} \boldsymbol{dist}(\hat{y}^{1}_{t},y^{1}_{t})
\end{equation}
\begin{equation}
    J_2(\theta)=\frac{1}{N}\displaystyle \sum_{n=1}^{N} \sum_{t=1}^{t=T} \boldsymbol{dist}(\hat{y}^{2}_{t},y^{2}_{t})
\end{equation}
$\boldsymbol{dist}(\hat{y},y)$ is a function that calculate the distance between $\hat{y}$ and $y$ such as Euclidean distance. Given the objective function for the first and second RNN as $J_{1}(\theta)$ and $J_{2}(\theta)$, respectively, the final objective function is defined as:
\begin{equation}
    J(\theta) = \alpha J_1(\theta) + (1-\alpha) J_2(\theta)
\end{equation}
By means of $\alpha$, we can prioritize between the first and the second objectives in the JRNN. The complete description about the JRNN could be found in~\cite{zhang}.

\subsection{Data Augmentation}
To increase the size of our training examples, we used a data augmentation method based on lexical dictionary. Given a set of training examples $Q=\{q_1, q_2, q_3, \dots, q_i\}$ and a lexical database T, we have to define the number of new training examples that should be generated from each training example, i.e. $n$ and the number of words that will be replaced for each text, i.e. $m$ in order to use our method. The algorithm~\ref{algo:aug} describes our augmentation method.

\begin{algorithm}
\KwIn{$Q, T, n, m, StopWord$}
\KwOut{The new training set $\hat{Q}$}
$\hat{Q} \gets Q$\;
\For{$q_i$ in Q}{
    $wordIndex$\;
    $synsetIndex$\;
    $index = 0$\;
    $\hat{Q}$\;
    \For{word $t_j$ in $q_i$}{
        \If{not ($t_j \in StopWord$)}{
            $wordIndex[index] \gets j$\;
            $synsetIndex[index] \gets T.index(T.getSynsetList(t_j))$\;
            $index++$\;
        }
    }
    $\hat{q} \gets q_i$\;
    \For{t in range n}{
        \eIf{$wordIndex.size \leq m$}{
            $replacedWords \gets range(0, wordIndex.size)$\;
        }{
            $replacedWords \gets 
                mRandom(0, wordIndex.size)$\;
        }
        \For{k in range RelacedWords.size}{
            $wordId = ReplacedWords[k]$\;
            $word \gets wordIndex[wordId]$\;
            $synonyms \gets random(synsetIndex[wordId])$\;
            $\hat{q}[word] \gets random(synonyms)$\;
        }
    }
    $\hat{Q}.append(\hat{q})$\;
}
\Return{$\hat{Q}$}\;
\caption{Our Data Augmentation Method}
\label{algo:aug}
\end{algorithm}

\subsection{Features}
\label{subsection:feature}
Zhang et al.~\cite{zhang} used only word embedding for the input in their JRNN. Unlike them, we used some additional linguistic features. In the following section, we describe our additional features, i.e. part of speech, named entity types and sentence dependency structure. These additional features are concatenated to word embedding feature. 

\subsubsection{Part of Speech Feature}
Part of speech or POS is a class where a word is assigned according to its syntactic role in a sentence such as adjective, verb, or noun. To identify it, we used a part of speech tagger. The idea is to use these categories as additional feature. 

POS are useful in this task because usually words that are part of keyphrases in a text belong to certain word categories such as noun, proper noun, and verb. In addition, some words in certain categories tend not to be a part of keyphrase such as preposition, adverb, and character symbol. 

For example, from the following text: ``ternyata di App AA sudah bisa beli paket internet Telkomsel sama Indosat ya, boleh juga", which means in English: ``it turns out that in App AA we have been able to buy Telkomsel internet packages and Indosat, alright then" has following keyphrases:
\begin{itemize}
    \item ``App AA'' tagged as proper noun,
    \item ``beli/buy'' tagged as verb intransitive,
    \item ``paket internet/internet packages'' tagged as noun,
    \item ``Telkomsel'' tagged as proper noun,
    \item ``Indosat'' tagged proper noun.
\end{itemize}
While some of the non-keyphrases are:
\begin{itemize}
    \item ``ternyata/turns out" tagged as adverb, 
    \item ``di/in" tagged as proposition,  
    \item ``sudah bisa/have been able to" tagged as TAME (tense, aspect, modality, and evidentiality).
\end{itemize}

We used INACL word categories~\cite{ayu1} as the guideline in POS tagging process. Our POS tagger model was trained with BiLSTM-CRF architecture~\cite{ma} and achieved F1 score of .9613 using Indonesian data consist of $61601$ train data and $10890$ test data in words. 

\subsubsection{Named Entity Feature}
Named entity recognition is one of the sub-tasks in information extraction that finds and classifies named entity mentions in an unstructured text into predefined categories such as the person names, organizations, locations, time, quantities, currency, etc.

Named entity or NE can be used to distinguish the type of entity that contains the same word such as the word ``java" in ``java programming language and ``Java island". Moreover, NE is useful in keyphrase extraction since some of keyphrases usually belong to some NE types. For example, the following text: ``tetapi App AA itu harus daftar dahulu ya di customer service bank" that means in English: ``but to have App AA it is necessary to register in bank customer service" has a keyphrase ``App AA" that belongs to name entity type organization. 

We used NE model based on BiLSTM-CRF architecture proposed by Hoesen and Purwarianti~\cite{hoesen}. This method achieved $.9210$ in F1 score using a dataset consisting of $179230$ train examples and $36680$ test (in words).

\subsubsection{Sentence Dependency Structure Feature}
A dependency structure of a sentence could be obtained by using a dependency parser. Dependency parser is a task to parse the syntactic structure of a sentence based on words' function (i.e. part of speech) and connectedness. Dependency parser focuses on determining word-based structures.  Therefore, we could get the syntactic and semantic contexts from a sentence. 

We used dependency structure or DS feature since keyphrase in a text often has dependency relationship with some word with certain POS tag.  For example, the following text: ``registrasi App AA lewat atm saja kan ya enggak perlu ke customer service", which is in English translated as: ``App AA registration can be done simply through ATM, does not need to go to the customer service" has a keyphrase ``customer service". According to dependency parser, words in this keyphrase have a possessive nominal modifier relationship from ``customer" to ``service". Hence, ``customer" is the parent of ``service".

We used dependency parser proposed by Rahman~\cite{rahman}. Their method is based on Bi-LSTM for generating feature vectors per token and used multilayer perceptron for parsing model. The performance of their method was .9613 in accuracy using dataset consisting of $17525$ training examples and $5274$ test data in words. Since dependency parser requires a POS tagger, they used the POS tag guideline from Purwarianti et al.~\cite{ayu1}.

\section{Experiments}
\label{sec:experiment}

\subsection{Data Construction}
In this experiment, we targeted texts from Twitter about banking products and services. To produce the dataset, we have two data annotators who manually labeled the data.

We have $1000$ training examples and $247$ test data for the evaluation. A phrase in a text is considered a keyphrase if it contains important information of the text. Different from Zhang et al.~\cite{zhang} which for each text has only one keyphrase, in our dataset, each text may contain multiple keyphrases because in our annotation effort, we found that important information of a text can be located separately within the sentence. It means that a keyphrase is often not enough to represent the important ideas of a text. 

For example, the following text: ``Bank A jelek amat 2 hari ini susah banget mau cek mutasi doang" that in English is translated as ``Bank A really sucks these 2 days it is really hard to even check balance" has following keyphrases: ``Bank A", ``jelek/sucks", ``susah/hard", ``cek mutasi/check balance". ``Bank A" is considered as keyphrase because it is the main object of the complaint, "jelek/sucks" is important because it shows user’s sentiment towards ``Bank A", ``susah/hard" is important because it is the main problem that user was experienced with, and ``cek mutasi/check balance" is considered a keyphrase because it is the source of the problem. Each keyphrase represents important idea of a text, which is very crucial to the text and one of them should not be missed. Additionally, they all are located separately in the text. 

After two annotators finished labelling the dataset, we have one senior annotator as a quality assurance to check the entire dataset and make correction if there are labeling errors. After the dataset evaluation was finished by the senior annotator, we found that the senior annotator agreed with $90\%$ of the data which are labelled manually by the two annotators. Hence, we concluded that this data is good for evaluating our method. This also shows that our manually labeled dataset is fit as ground-truth for this keyphrase extraction task. Table 1 shows the statistic of our dataset. 

\begin{table}[t!]
\caption{\label{font-table} Statistics of our dataset}
\label{tab:statistic}
\begin{center}
\begin{tabular}{|l|r|r|r|r|}
\hline \bf Attribute & \bf Train & \bf Test \\ \hline
 Total Data & $1000$ & $247$ \\ 
 Total Keyphrase & $4198$ & $1289$ \\
 Average Keyphrase & $4$ & $5$ \\
 Total Words & $11947$ & $3708$ \\
 Total Class 0 & $6172$ & $2216$ \\
 Total Class 1 & $4019$ & $983$ \\
 Total Class 2 & $1756$ & $507$ \\
\hline
\end{tabular}
\end{center}
\end{table}

\subsection{Experiment Configurations}
To perform the experiment, from total $1000$ training data and $247$ test data, we split $10\%$ of the training data as the validation set. Hence, we have $900$ tweets as the training set, $100$ texts as the validation set, and $247$ as test data for the evaluation. 

For the evaluation method, we used precision (P), recall (R),  F1-score (F1) and Accuracy (Acc). These metrics are calculated based on true positive (TP), true negative (TN), false positive (FP), false negative (FN) counted during the evaluation. TP is the number of words labelled as 1 or 2 that are correctly classified in the evaluation. TN is the number of words in the class of 0 which are correctly predicted. FP is the number of words in the class of 1 or 2 that are classified as non-keyphrase. FN is the number of words that are labelled as 0 classified as 1 or 2. 

R is the ratio of TP and the total number of words in the class of 1 or 2. P is the ratio of TP and the total number of words that are classified as 1 or 2 in the evaluation. F1 is the harmonic combination between P and R. A is calculated as TP and TN divided by the total number of words in the dataset. 

As we have mentioned before in the sub-section~\ref{subsection:feature}, we also used a word embedding representation as the input to the JRNN. The word embedding we used were pre-trained vectors on indonesian news articles containing $253849$ words. We used a skip-gram model proposed by Mikolov et al.~\cite{mikolov} with $100$ dimensions.

The default parameters of our method are as follows: the learning rate is 0.1, the window size is 3, the number of neurons in the first and second hidden layer are $300$, $\alpha$ is $.5$. For the data augmentation, for each training example we generate three new examples ($n$) with three word replacement in each generated example ($m$). We only applied the data augmentation on the train examples. Therefore, we could avoid the possibility of over-fitting with the test data on our model. In addition, we could compare the performance of our method with data augmentation or not. Hence, the effect of data augmentation could be observed.

\subsection{Methods for Comparison}
We compared our method with several existing methods that belong to unsupervised and supervised method. For the unsupervised method, we used RAKE~\cite{rose}. While for the supervised one, we used RNN, LSTM, and the original JRNN with 5 classes (JRNN5) proposed by Zhang et al.~\cite{zhang}. For the supervised methods, we used our word embedding as their input. 

\textbf{RNN}: A recurrent neural network (RNN) is a neural network where the links between cell units form a directed cycle graph. This situation creates an internal state of the network that shows dynamic temporal behavior. 

\textbf{LSTM}: Long short-term memory (LSTM) is a variant of RNNs. Unlike the traditional ones, an LSTM network is appropriate to learn patterns to classify, process, and predict time series in which there are time lags between the important events~\cite{hochreiter}. 

\textbf{RAKE}: Rapid Automatic Keyword Extraction (RAKE) is a domain independent keyword extraction algorithm that determines keyphrases in a text by analyzing the word frequency and its co-occurrence with other words in the text~\cite{rose}.

\subsection{Experiment Results}
Table 2 shows the performances of our method and the baseline methods on our dataset. Among the baseline methods, LSTM-WE outperformed the other ones in terms of F1 and Acc. RAKE performed the worst because this method only depends on stopwords and statistical information in terms of all evaluation methods. The second worst in terms of Acc is achieved by JRRN5 proposed by Zhang et al.~\cite{zhang}, it was because five classes made keyphrase extraction task much harder. Although more classes means more detailed results but in our case with only $1000$ training data it showed that this approach did not give the best solution.

Among the proposed methods, our method that combined all features and performed data augmentation achieved the best Acc and P. While the best of both F1 and R were achieved by our method JRNN3-WE-POS. Both methods (all features and data augmented) in terms of their best scores outperformed all the baseline methods including JRNN5 proposed by Zhang et al.~\cite{zhang}. Our method that combined all features and data augmentation performed $.1800$ and $.0923$ higher that JRNN5 in terms of Acc and R, respectively. 

In general, our method JRNN3-WE-POS achieved the best performance in all evaluation methods compared to the baselines, JRNN3-WE-NE and JRNN3-WE-DS. This shows that POS features are the most important aspect in this task compared to NE and DS. Since the performance of our method that used NE is better than DS, NE is the second important feature in this task. Although by adding DS to JRNN3-WE decreased the performance of our method in terms of R and F1, this method has better performance in terms of P and Acc. The above findings suggests that our additional features are useful for extracting keyphrases. 

In regards of augmentation, our method JRNN3-WE-POS-DS-Augmentation achieved the best P and Acc. The increase in precision shows that the data augmentation gave synonym variations to keyphrases extracted during training, which made the model more flexible while predicting testing data. Although the use of data augmentation method increased P, it lowered R as we observe by comparing it to JRRN3-WE-POS-NE-DS. This happened due to the data variations in the augmented training data, causing the model to be more precise in predicting certain data but sacrificing the ability of the model to predict certain test data containing new keyphrases that do not have variations in the augmented training data. 

Table 3 shows our method with various values of $\alpha$. According to these results, we found that the best $\alpha$ is $.3$ since the method with this configuration achieved the best Acc and F1. It means that we have to prioritize the objective for learning whether a word is non-keyphrase, the beginning of keyphrase, or the tail of keyphrase. It may be due to our dataset that contains multiple keyphrases from a single text.  

On the other hand, the highest precision achieved by $\alpha=.5$, which showed that sharing the same weight provides maximum P but sacrificing R. It means that the method was strictly selecting keyphrases from a text. The highest recall achieved by $.9$, which showed that prioritizing keyword tagging task over keyphrase tagging task will gave higher R but sacrificing P. Based on the above findings regarding the $\alpha$, we suggest to use $\alpha=.3$. 

\begin{table*}[t!]
\caption{\label{font-table} The effectiveness of our method compare to the baselines}
\label{tab:experimentall}
\begin{center}
\begin{tabular}{|l|r|r|r|r|}
\hline \bf Method & \bf P & \bf R & \bf F1 & \bf Acc \\ \hline
 \textit{Baselines} &  &  &  &  \\
 RAKE~\cite{rose} & $.4454$ & $.7557$ & $.5604$ & $.6112$ \\
 RNN-WE & $.6665$ & $.8883$ & $.7616$ & $.7839$ \\
 LSTM-WE & $.6711$ & $.9069$ & $.7714$ & $.7878$  \\
 JRNN5-WE~\cite{zhang} & $.6235$ & $.8685$ & $.7258$ & $.7793$ \\ \hline
\textit{ Our methods} &  &  &  &  \\
 JRNN3-WE & $.6621$ & $.9000$ & $.7629$ & $.7962$ \\
 JRNN3-WE-POS & $.6759$ & $\boldsymbol{.9117}$ & $\boldsymbol{.7763}$ & $.9066$ \\
 JRNN3-WE-NE & $.6709$ & $.9087$ & $.7719$ & $.9023$ \\ 
 JRNN3-WE-DS & $.6973$ & $.7882$ & $.7400$ & $.9336$ \\
 JRNN3-WE-POS-NE  & $.6781$ & $.8883$ & $.7691$ & $.9360$ \\
 JRNN3-WE-POS-DS & $.6910$ & $.7965$ & $.7400$ & $.9498$ \\
 JRNN3-WE-NE-DS & $.6991$ & $.7958$ & $.7443$ & $.9502$ \\
 JRNN3-WE-NE-POS-DS & $.6955$ & $.7752$ & $.7332$ & $.9591$ \\
 JRNN3-WE-NE-POS-DS-Augmentation & $\boldsymbol{.7158}$ & $.7272$ & $.7215$ & $\boldsymbol{.9593}$ \\
\hline
\end{tabular}
\end{center}
\end{table*}

\begin{table}[t!]
\caption{\label{font-table} The effectiveness of our method with various $\alpha$}
\label{tab:alpha}
\begin{center}
\begin{tabular}{|l|r|r|r|r|}
\hline \bf $\alpha$ & \bf P & \bf R & \bf F1 & \bf Acc \\ \hline
 $.1$ & $.6852$ & $.8061$ & $.7407$ & $.9591$ \\ 
 $.3$ & $.6903$ & $.8049$ & $\boldsymbol{.7432}$ & $\boldsymbol{.9597}$ \\
 $.5$ & $\boldsymbol{.6955}$ & $.7752$ & $.7332$ & $.9591$ \\
 $.7$ & $.6906$ & $.7888$ & $.7364$ & $.9591$ \\
 $.9$ & $.6710$ & $\boldsymbol{.8069}$ & $.7327$ & $.9582$ \\
\hline
\end{tabular}
\end{center}
\end{table}

\section{Conclusion and Future Work}
\label{sec:conclusion}
In this work, we proposed a novel combination of Joint Layer Recurrent Neural Network (JRNN). Different from the original JRNN, we simplified the learning prediction task by reducing the number of classes. Additionally, we also proposed to use linguistic features in the JRNN, i.e. part of speech, named entity types, and sentence structure dependency. Since the number of training examples in our dataset is small, we also proposed to use a data augmentation method to increase its size. 

We evaluated the proposed method on $247$ test data consist of Indonesian texts from Twitter on banking domain which labeled manually by human annotators. Our experiments showed that the proposed method outperformed the baseline methods. We found that part of speech is one of the important features in JRNN for extracting keyphrases. We found that the best $\alpha$ is $.3$. Our experiments also suggest that our data augmentation improved the performance of our method in terms of P and Acc. 

For the future works, it is possible to improve the performance of the data augmentation by using certain methods to choose the synset from lexical dictionary. Additionally, when choosing word from the synset, we could also consider its part of speech or named entity type. To improve our keyphrase extraction method in general, we could use more linguistic features, like the information from constituent parser, co-reference resolution, or positional feature of word relative to sentence.

\end{document}